\documentclass[a4paper,twoside]{article}

\usepackage{slashbox}
\usepackage{natbib}
\usepackage{epsfig}
\usepackage{subfigure}
\usepackage{calc}
\usepackage{amssymb}
\usepackage{amstext}
\usepackage{amsmath}
\usepackage{amsthm}
\usepackage{multicol}
\usepackage{pslatex}
\usepackage{apalike}

\usepackage{epstopdf}
\usepackage{mathtools}
\usepackage[english]{babel}
\usepackage[utf8]{inputenc}
\graphicspath{{imgs/}{../imgs/}}
\usepackage[hidelinks]{hyperref}
\hypersetup{
  colorlinks = true,
  urlcolor   = blue,
  linkcolor  = blue,
  citecolor  = red
}

\usepackage{SCITEPRESS}

\begin{document}

\title{The Optional Prisoner's Dilemma in a Spatial Environment: Coevolving Game Strategy and Link Weights
\thanks{The final publication will be available in the Proceedings of the 8th International Joint Conference on Computational Intelligence. \url{http://www.ecta.ijcci.org}}
}

\author{
    \authorname{Marcos Cardinot\sup{1}, Colm O'Riordan\sup{1} and Josephine Griffith\sup{1}}
    \affiliation{\sup{1}Information Technology, National University of Ireland, Galway, Ireland}
    \email{\{marcos.cardinot, colm.oriordan, josephine.griffith\}@nuigalway.ie}
}


\abstract{
In this paper, the Optional Prisoner's Dilemma game in a spatial environment,
with coevolutionary rules for both the strategy and network links between
agents, is studied. Using a Monte Carlo simulation approach, a number of
experiments are performed to identify favourable configurations of the
environment for the emergence of cooperation in adverse scenarios. Results show
that abstainers play a key role in the protection of cooperators against
exploitation from defectors. Scenarios of cyclic competition and of full
dominance of cooperation are also observed. This work provides insights towards
gaining an in-depth understanding of the emergence of cooperative behaviour in
real-world systems.
}

\onecolumn \maketitle \normalsize \vfill

\section{\uppercase{Introduction}}
\label{sec:introduction}
\noindent
Evolutionary game theory in spatial environments has attracted much interest
from researchers who seek to understand cooperative behaviour among rational
individuals in complex environments. Many models have considered the scenarios
where participant’s interactions are constrained by particular graph
topologies, such as lattices \citep{Hauert2002,Nowak1992}, small-world graphs
\citep{Chen2008,Fu2007}, scale-free graphs \citep{Szolnoki2016,Xia2015} and,
bipartite graphs \citep{Gomez2011}. It has been shown that the spatial
organisation of strategies on these topologies affects the evolution of
cooperation \citep{Cardinot2016}.

The Prisoner's Dilemma (PD) game remains one of the most studied games in
evolutionary game theory as it provides a simple and powerful framework to
illustrate the conflicts in the formation of cooperation. In addition, some
extensions of the PD game, such as the Optional Prisoner's Dilemma game, have
been studied in an effort to investigate how levels of cooperation can be
increased. In the Optional PD game, participants are afforded a third option
--- that of abstaining and not playing and thus obtaining the loner's payoff ($L$).
Incorporating this concept of abstinence leads to a three-strategy game where
participants can choose to cooperate, defect or abstain from a game
interaction.

The vast majority of the spatial models in previous work have used static and
unweighted networks. However, in many social scenarios that we wish to model,
such as social networks and real biological networks, the number of
individuals, their connections and environment are often dynamic. Thus, recent
studies have also investigated the effects of evolutionary games played on
dynamically weighted networks
\citep{Huang2015,Wang2014,Cao2011,Szolnoki2009,Zimmermann2004} where it has
been shown that the coevolution of both networks and game strategies can play
a key role in resolving social dilemmas in a more realistic scenario.

In this paper we adopt a coevolutionary spatial model in which both the game
strategies and the link weights between agents evolve over time. The
interaction between agents is described by an Optional Prisoner's Dilemma game.
Previous research on spatial games has shown that when the temptation to
defect is high, defection is the dominant strategy in most cases. We believe
that the combination of both optional games and coevolutionary rules can help
in the emergence of cooperation in a wider range of scenarios.

The aims of the work are, given an Optional Prisoner's Dilemma game in a
spatial environment, where links between agents can be evolved, to understand
the effect of varying the:
\begin{itemize}
    \item value of the link weight amplitude (the ratio $\Delta/\delta$).
    \item value of the loner's payoff ($L$).
    \item value of the temptation to defect ($T$).
\end{itemize}

By investigating the effect of these parameters, we aim to explore the impact
of the link update rules and to investigate the evolution of cooperation when
abstainers are present in the population.

Although some work has considered coevolving link weights when considering the
Prisoner's Dilemma in a spatial environment, to our knowledge the investigation
of an Optional Prisoner's Dilemma game on a spatial environment, where both the
strategies and link weights are evolved, has not been studied to date.

The results show that cooperation emerges even in extremely adverse scenarios
where the temptation of defection is almost at its maximum. It can be observed
that the presence of the abstainers are fundamental in protecting cooperators
from invasion. In general, it is shown that, when the coevolutionary rules are
used, cooperators do much better, being also able to dominate the whole
population in many cases. Moreover, for some settings, we also observe
interesting phenomena of cyclic competition between the three strategies, in
which abstainers invade defectors, defectors invade cooperators and cooperators
invade abstainers.

The paper outline is as follows: Section~\ref{sec:related} presents a brief
overview of the previous work in both spatial evolutionary game theory with
dynamic networks and in the Optional Prisoner's Dilemma game.
Section~\ref{sec:methodology} gives an overview of the methodology employed,
outlining the Optional Prisoner's Dilemma payoff matrix, the coevolutionary
model used (Monte Carlo simulation), the strategy and link weight update rules,
and the parameter values that are varied in order to explore the effect of
coevolving both strategies and link weights.  Section~\ref{sec:results}
features the results.  Finally, Section~\ref{sec:conclusion} summarizes the
main conclusions and outlines future work.

\section{\uppercase{Related Work}}
\label{sec:related}
\noindent
The use of coevolutionary rules constitute a new trend in evolutionary game
theory.  These rules were first introduced by \citet{Zimmermann2001}, who
proposed a model in which agents can adapt their neighbourhood during a
dynamical evolution of game strategy and graph topology. Their model uses
computer simulations to implement two rules: firstly, agents playing the
Prisoner's Dilemma game update their strategy (cooperate or defect) by
imitating the strategy of an agent  in their neighbourhood with a higher
payoff; and secondly, the network is updated by allowing defectors to break
their connection with other defectors and replace the connection with a
connection to a new neighbour selected randomly from the whole network.
Results show that such an adaptation of the network is responsible for an
increase in cooperation.

In fact, as stated by \citet{Perc2010}, the spatial coevolutionary game is a
natural upgrade of the traditional spatial evolutionary game initially proposed
by \citet{Nowak1992}, who considered static and unweighted networks in which
each individual can interact only with its immediate neighbours. In general, it
has been shown that coevolving the spatial structure can promote the emergence
of cooperation in many scenarios \citep{Wang2014,Cao2011}, but the
understanding of cooperative behaviour is still one of the central issues in
evolutionary game theory.

\citet{Szolnoki2009} proposed a study of the impact of coevolutionary rules on
the spatial version of three different games, i.e., the Prisoner's Dilemma, the
Snow Drift and the Stag Hunt game. They introduce the concept of a teaching
activity, which quantifies the ability of each agent to enforce its strategy on
the opponent. It means that agents with higher teaching activity are more
likely to reproduce than those with a low teaching activity.  Differing from
previous research \citep{Zimmermann2004,Zimmermann2001}, they also consider
coevolution affecting either only the defectors or only the cooperators. They
discuss that, in both cases and irrespective of the applied game, their
coevolutionary model is much more beneficial to the cooperators than that of
the traditional model.

\citet{Huang2015} present a new model for the coevolution of game strategy and
link weight. They consider a population of $100 \times 100$ agents arranged on
a regular lattice network which is evolved through a Monte Carlo simulation.
An agent's interaction is described by the classical Prisoner's Dilemma with a
normalized payoff matrix. A new parameter, $\Delta/\delta$, is defined as the
link weight amplitude and is calculated as the ratio of $\Delta/\delta$. They found that some values of $\Delta/\delta$ can
provide the best environment for the evolution of cooperation. They also found
that their coevolutionary model can promote cooperation efficiently even when
the temptation of defection is high.

In addition to investigations of the classical Prisoner's Dilemma on spatial
environments, some extensions of this game have also been explored as a means
to favour the emergence of cooperative behaviour. For instance, the Optional
Prisoner's Dilemma game, which introduces the concept of abstinence, has been
studied since \citet{Batali1995}. In their work, they proposed the opt-out or
``loner's'' strategy in which agents could choose to abstain from playing the
game, as a third option, in order to avoid cooperating with known defectors.
There have been a number of recent studies exploring this type of game
\citep{Xia2015,Ghang2015,Olejarz2015,Jeong2014,Hauert2008}.
\citet{Cardinot2016} discuss that, with the introduction of abstainers, it is
possible to observe new phenomena and a larger range of scenarios where
cooperators can be robust to invasion by defectors and can dominate.

However, the inclusion of optional games with coevolutionary rules has not
been studied yet. Therefore, our work aims to combine both of these trends in
evolutionary game theory in order to identify favourable configurations for the
emergence of cooperation in adverse scenarios, where, for example, the
temptation to defect is very high.

\section{\uppercase{Methodology}}
\label{sec:methodology}
\noindent
The goal of the experiments outlined in this section is to investigate the
environmental settings when coevolution of both strategy and link weights of
the Optional Prisoner's Dilemma on a weighted network takes place.

Firstly, the Optional Prisoner's Dilemma (PD) game will be described; secondly,
the spatial environment is described; thirdly, the coevolutionary rules for
both the strategy and link weights are described and finally, the experimental
set-up is outlined.

In the classical version of the Prisoner's Dilemma, two agents can choose
either cooperation or defection. Hence, there are four payoffs associated with
each pairwise interaction between the two agents. In consonance with common
practice \citep{Huang2015,Nowak1992}, payoffs are characterized by the reward
for mutual cooperation ($R=1$), punishment for mutual defection ($P=0$),
sucker's payoff ($S=0$) and temptation to defect ($T=b$, where $1<b<2$).
Note that this parametrization refers to the weak version of the Prisoner's
Dilemma game, where $P$ can be equal to $S$ without destroying the nature of
the dilemma. In this way, $T > R > P \ge S$ maintains the dilemma.

The extended version of the PD game presented in this paper includes the
concept of abstinence, in which agents can not only cooperate ($C$) or defect
($D$) but can also choose to abstain ($A$) from a game interaction, obtaining
the loner's payoff ($L=l$) which is awarded to both players if one or both
abstain. As defined in other studies \citep{Cardinot2016,Hauert2002},
abstainers receive a payoff greater than $P$ and less than $R$ (i.e., $P<L<R$).
Thus, considering the normalized payoff matrix adopted, $0<l<1$. The payoff
matrix and the associated values are illustrated in Tables \ref{tab:payoff1}
and \ref{tab:payoff2}.

\begin{table}[htb]
    \centering
    \caption{The Optional Prisoner's Dilemma game matrix.}
    \label{tab:payoff1}
    \begin{tabular}{c c | c | c}
            & {\bf C} & {\bf D} & {\bf A} \\
        \cline{2-4}
        {\bf C}  & \multicolumn{1}{|c|}{\backslashbox{R}{R}}
                 & \backslashbox{S}{T}
                 & \multicolumn{1}{c|}{\backslashbox{L}{L}} \\
        \cline{1-4}
        {\bf D}  & \multicolumn{1}{|c|}{\backslashbox{T}{S}}
                 & \backslashbox{P}{P}
                 & \multicolumn{1}{c|}{\backslashbox{L}{L}} \\
        \cline{1-4}
        {\bf A}  & \multicolumn{1}{|c|}{\backslashbox{L}{L}}
                 & \backslashbox{L}{L}
                 & \multicolumn{1}{c|}{\backslashbox{L}{L}} \\
        \cline{2-4}
    \end{tabular}
\end{table}

\begin{table}[htb]
    \centering
    \caption{Payoff values.}
    \label{tab:payoff2}
    \setlength{\tabcolsep}{8pt}
    \begin{tabular}{l|c}
        {\bf Payoff} & {\bf Value} \\
        \hline
        {Temptation to defect (T)}              & $]1,2[$     \\
        {Reward for mutual cooperation (R)}     & $1$         \\
        {Punishment for mutual defection (P)}   & $0$         \\
        {Sucker's payoff (S)}                   & $0$         \\
        {Loner's payoff (L)}                    & $]0,1[$     \\
    \end{tabular}
\end{table}

In these experiments, the following parameters are used: a $100 \times 100$
~($N=100^2$) regular lattice grid with periodic boundary conditions is created
and fully populated with agents, which can play with their eight immediate
neighbours (Moore neighbourhood). We adopt an unbiased environment in which
initially each agent is designated as a cooperator ($C$), defector ($D$) or
abstainer ($A$) with equal probability. Also, each edge linking agents has the
same weight $w=1$, which will adaptively change in accordance with the
interaction.

Monte Carlo methods are used to perform the Optional Prisoner’s Dilemma game.
In one Monte Carlo (MC) step, each player is selected once on average. This
means that one MC step comprises $N$ inner steps where the following
calculations and updates occur:

\begin{itemize}
    \item Select an agent ($x$) at random from the population.

    \item Calculate the utility $u_{xy}$ of each interaction of $x$ with its
        eight neighbours (each neighbour represented as agent $y$) as follows:
        \begin{equation}
            u_{xy} = w_{xy} P_{xy},
        \end{equation}
        where $w_{xy}$ is the edge weight between agents $x$ and $y$, and
        $P_{xy}$ corresponds to the payoff obtained by agent $x$ on playing the
        game with agent $y$.

    \item Calculate $U_x$ the accumulated utility of $x$, that is:
        \begin{equation}
            U_x = \sum_{y \in \Omega_x}u_{xy},
        \end{equation}
        where $\Omega_x$ denotes the set of neighbours of the agent $x$.

    \item In order to update the link weights, $w_{xy}$ between agents, compare
        the values of $u_{xy}$ and the average accumulated utility
        (i.e., $\bar{U_x}=U_x/8$) as follows:
        \begin{equation}
            w_{xy} =
            \begin{dcases*}
                w_{xy} + \Delta  & if $u_{xy} > \bar{U_x}$ \\
                w_{xy} - \Delta  & if $u_{xy} < \bar{U_x}$ \\
                w_{xy}           & otherwise
            \end{dcases*},
        \end{equation}
        where $\Delta$ is a constant such that $0 \le \Delta/\delta \le 1$.

    \item In line with previous research \citep{Huang2015,Wang2014}, $w_{xy}$
        is adjusted to be within the range of $1-\delta$ to $1+\delta$, where
        $\delta$ ($0 < \delta \le 1$) defines the weight heterogeneity. Note
        that when $\Delta$ or $\delta$ are equal to $0$, the link weight keeps
        constant (${w=1}$), which results in the traditional scenario where
        only the strategies evolve.

    \item In order to update the strategy of $x$, the accumulated utility $U_x$
        is recalculated (based on the new link weights) and compared with the
        accumulated utility of one randomly selected neighbour ($U_y$).
        If $U_y>U_x$, agent $x$ will copy the strategy of agent $y$ with a
        probability proportional to the utility difference
        (Equation~\ref{eq:prob}), otherwise, agent $x$ will keep its strategy
        for the next step.
        \begin{equation}
            \label{eq:prob}
            p(s_x=s_y) = \frac{U_y-U_x}{8(T-P)},
        \end{equation}
        where $T$ is the temptation to defect and $P$ is the punishment for
        mutual defection. This equation has been considered previously by
        \citet{Huang2015}.
\end{itemize}

Simulations are run for $10^5$ MC steps and the fraction of cooperation is
determined by calculating the average of the final 1000 MC steps. To alleviate
the effect of randomness in the approach, the final results are obtained by
averaging 10 independent runs.

\noindent
The following scenarios are investigated:
\begin{itemize}
    \item Exploring the effect of the link update rules by varying the values of
        $\Delta/\delta$. Specifically, the value of the link weight amplitude
        $\Delta/\delta$ is varied for a range of fixed values of the loner's
        payoff (${l=[0.0,\ 1.0]}$), temptation to defect (${b=[1.0,\ 2.0]}$)
        and the weight heterogeneity (${\delta=(0.0,\ 1.0]}$).

    \item Investigating the evolution of cooperation when abstainers are present
        in the population.

    \item Considering snapshots of the evolution of the population over time at
        Monte Carlo steps of $0$, $45$, $1113$, and $10^5$.

    \item Investigating the relationship between $\Delta/\delta$, $b$ and $l$.
        Specifically, the values of $b$, $l$ and $\Delta/\delta$ are varied for
        a fixed value of $\delta$ ($\delta = 0.8$).
\end{itemize}

It is noteworthy that a wider range of values for $l$, $b$ and $\delta$ were
considered in our simulations, but for the sake of simplicity, we report only
the values of $l=\{0.0,\ 0.6\}$, $b=\{1.18,\ 1.34,\ 1.74,\ 1.90\}$ and
$\delta = \{0.2,\ 0.4,\ 0.8\}$, which are representative of the outcomes at
other values also.

\section{\uppercase{Results}}
\label{sec:results}
\noindent
In this section, we present some of the relevant experimental results of the
simulations of the Optional Prisoner's Dilemma game on the weighted network.

\subsection{Varying the Values of $\Delta/\delta$}

\begin{figure*}[htb]
    \centering
    {\epsfig{file=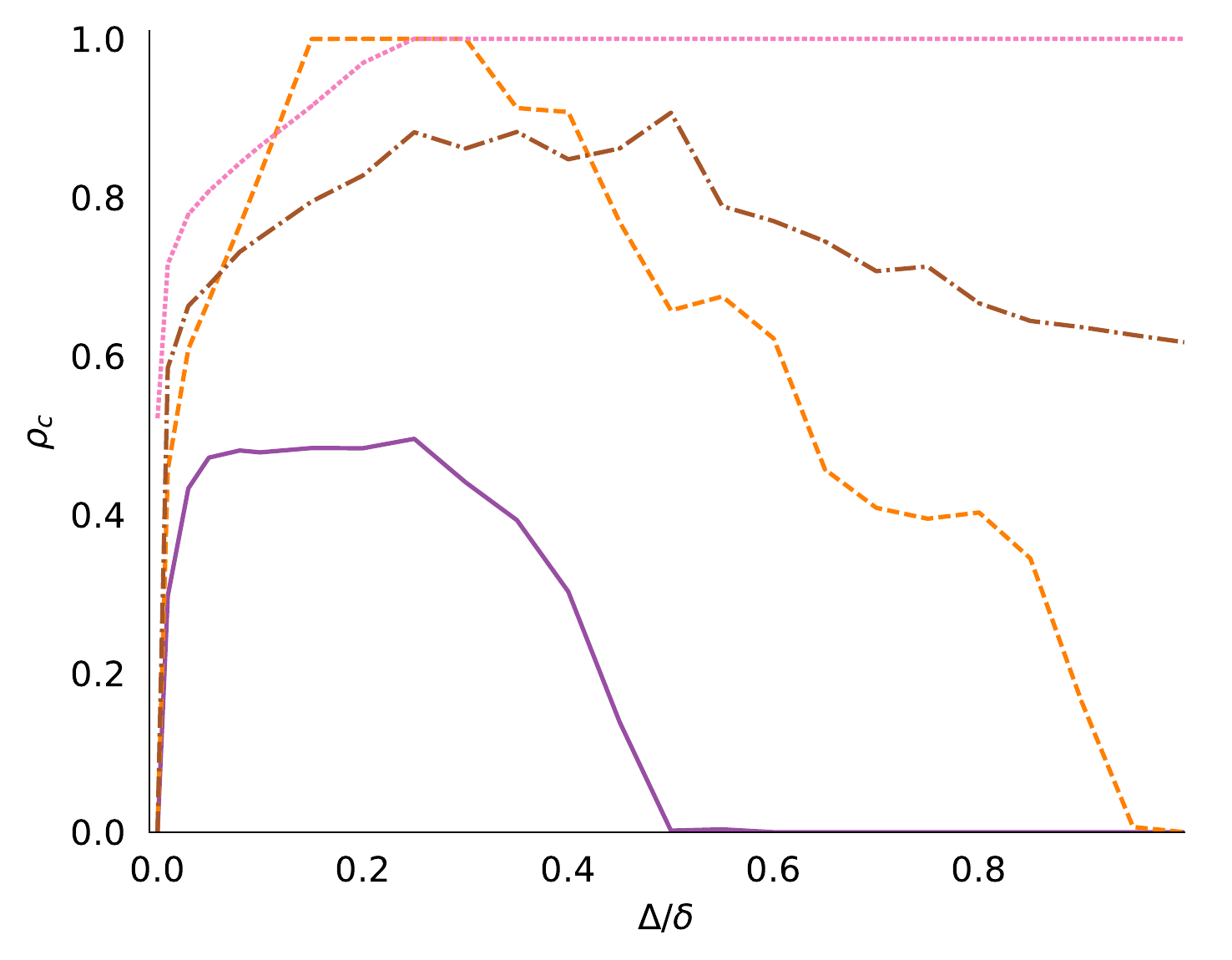, height=5.1cm}}
    {\epsfig{file=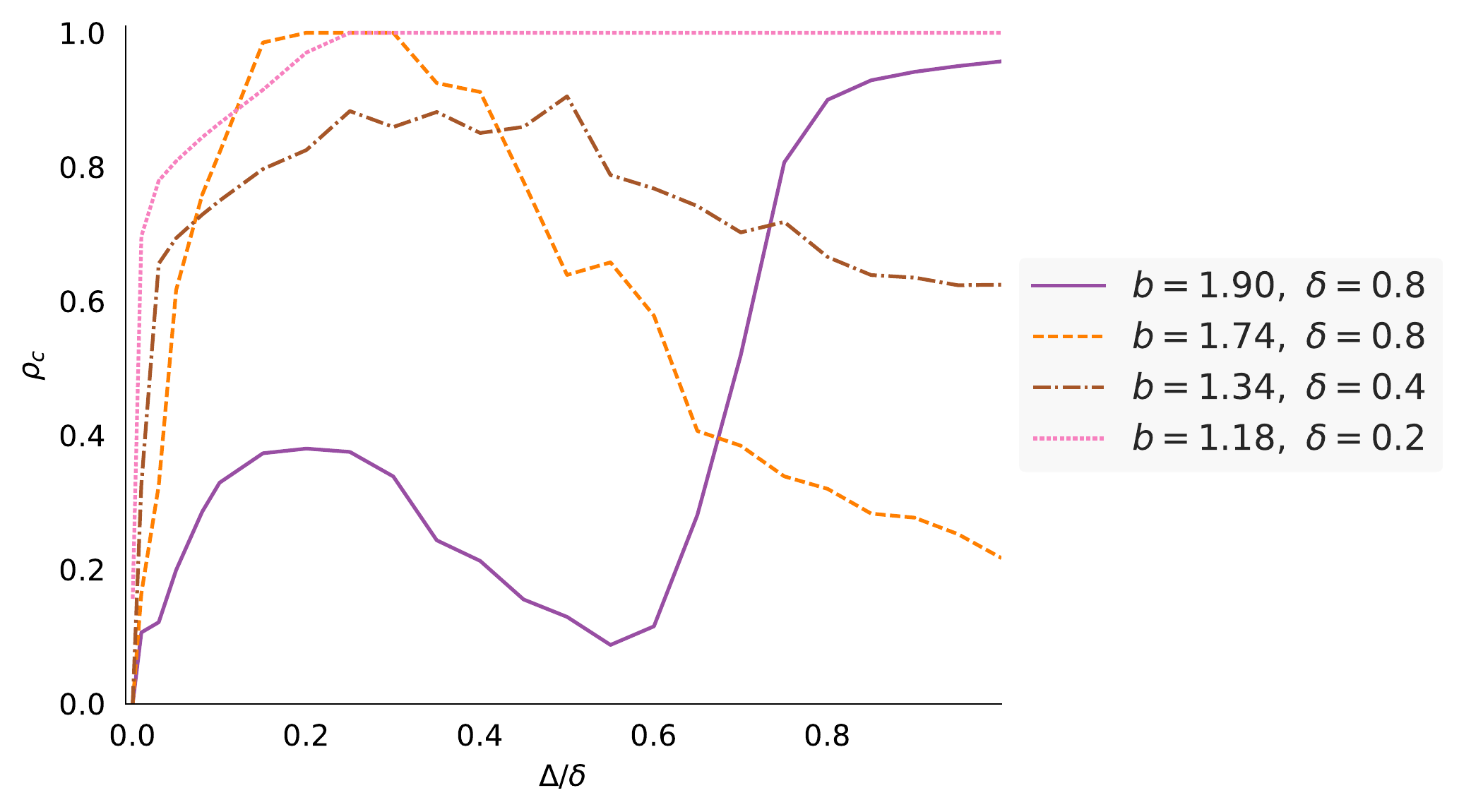, height=5.1cm}}
    \caption{
        Relationship between cooperation and link weight amplitude
        $\Delta/\delta$ when the loner's payoff ($l$) is equal to $0.0$ (left)
        and $0.6$ (right).
    }
    \label{fig:c_amp}
\end{figure*}

Figure~\ref{fig:c_amp} shows the impact of the coevolutionary model on the
emergence of cooperation when the link weight amplitude $\Delta/\delta$ varies
for a range of fixed values of the loner's payoff ($l$), temptation to defect
($b$) and weight heterogeneity ($\delta$). In this experiment, we observe that
when $l=0.0$, the outcomes of the coevolutionary model for the Optional
Prisoner's Dilemma game are very similar to those in the classical Prisoner's
Dilemma game \citep{Huang2015}. This result can be explained by the normalized
payoff matrix adopted in this work (Table~\ref{tab:payoff1}). Clearly, when
$l=0.0$, there is no advantage in abstaining from playing the game, thus agents
choose the option to cooperate or defect.

In cases where the temptation to defect is less than or equal to $1.34$ ($b \le 1.34$), it can be observed that
the level of cooperation does not seem to be affected by the increment of the
loner's payoff, except when the advantage in abstaining is very high, i.e., $l
\ge 0.8$. However, these results highlight that the presence of the abstainers
may protect cooperators from invasion. Moreover, the difference between the
traditional case ($\Delta/\delta=0.0$) for $l=\{0.0,\ 0.6\}$ and all other
values of $\Delta/\delta$ is strong evidence that our coevolutionary model is
very advantageous  to the promotion of cooperative behaviour. Namely, when
$l=0.6$, in the traditional case with a static and unweighted network
($\Delta/\delta=0.0$), the cooperators have no chance of surviving; in this
scenario, when the temptation to defect $b$ is low, abstainers always dominate,
otherwise, when $b$ is high, defection is always the dominant strategy.
However, when the coevolutionary rules are used, cooperators do much better,
being also able to dominate the whole population in many cases.

\subsection{Presence of Abstainers}

In addition to the fact that the levels of cooperation are usually improved in the
coevolutionary model, as the value of the loner's payoff increases we also
observe newer phenomena.

In the classical Prisoner's Dilemma in this type of environment, when the
defector's payoff is very high (i.e., greater than $1.75$) defectors spread
quickly and dominate the environment. However, Figure~\ref{fig:c_amp} also
shows that, for some values of $l$, it is possible to reach high levels of
cooperation even when the temptation of defection $b$ is almost at its peak.

Therefore, abstainers seem to help the population to increase their fraction of
cooperation in many cases, but mainly in the case where the link weight
amplitude $\Delta/\delta$ is higher than $0.7$. This is usually a bad scenario
in the classical Prisoner's Dilemma game.

\subsection{Snapshots at Different Monte Carlo Steps}

\begin{figure*}[p]
    \centering
    {\epsfig{file=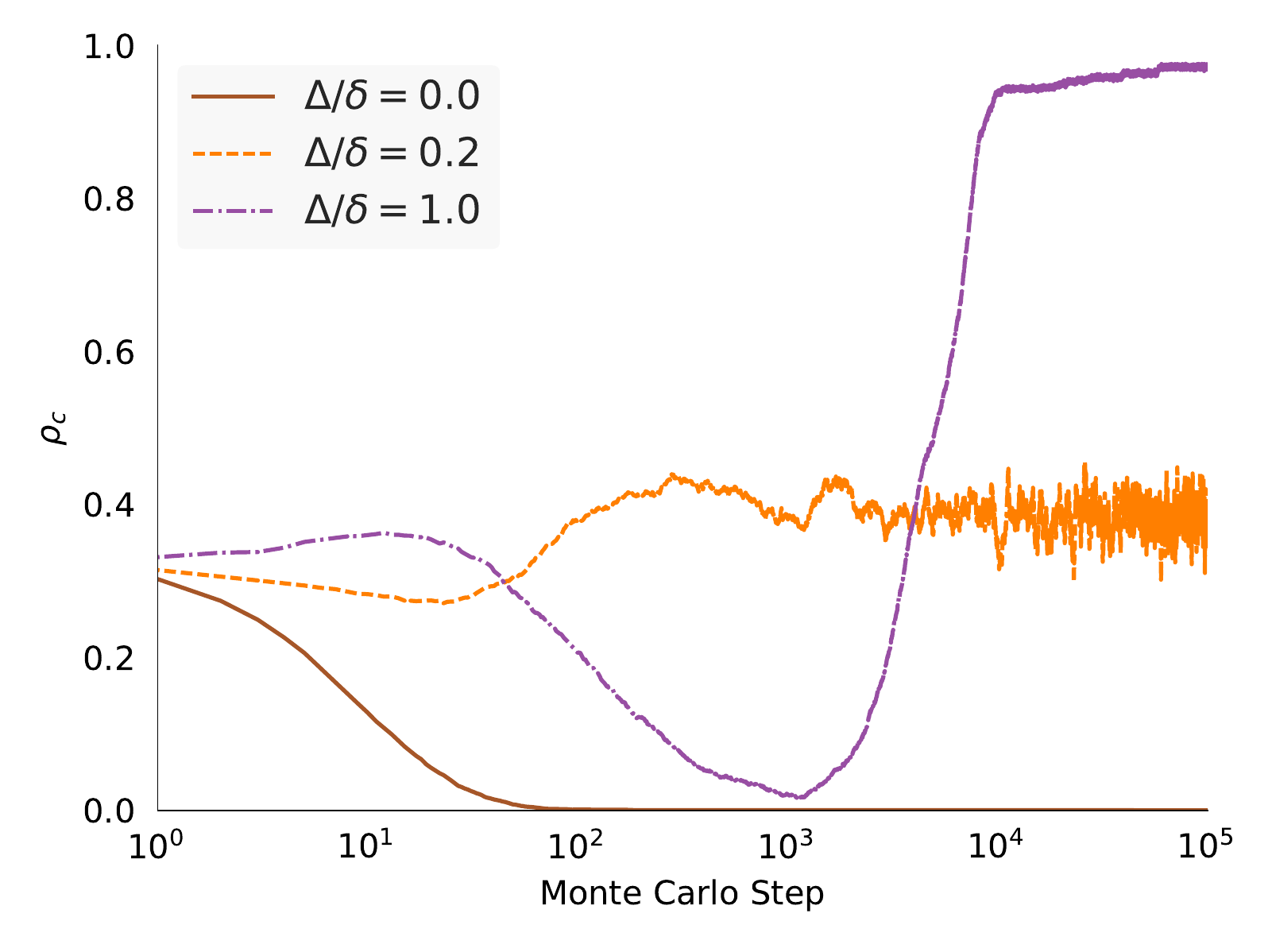, width=0.6\linewidth}}
    \caption{
        Progress of the fraction of cooperation $\rho_c$ during a Monte Carlo
        simulation for $b=1.9$, $l=0.6$ and $\delta=0.8$.
    }
    \label{fig:c_mcs}
\end{figure*}

\begin{figure*}[p]
    \centering
    {\epsfig{file=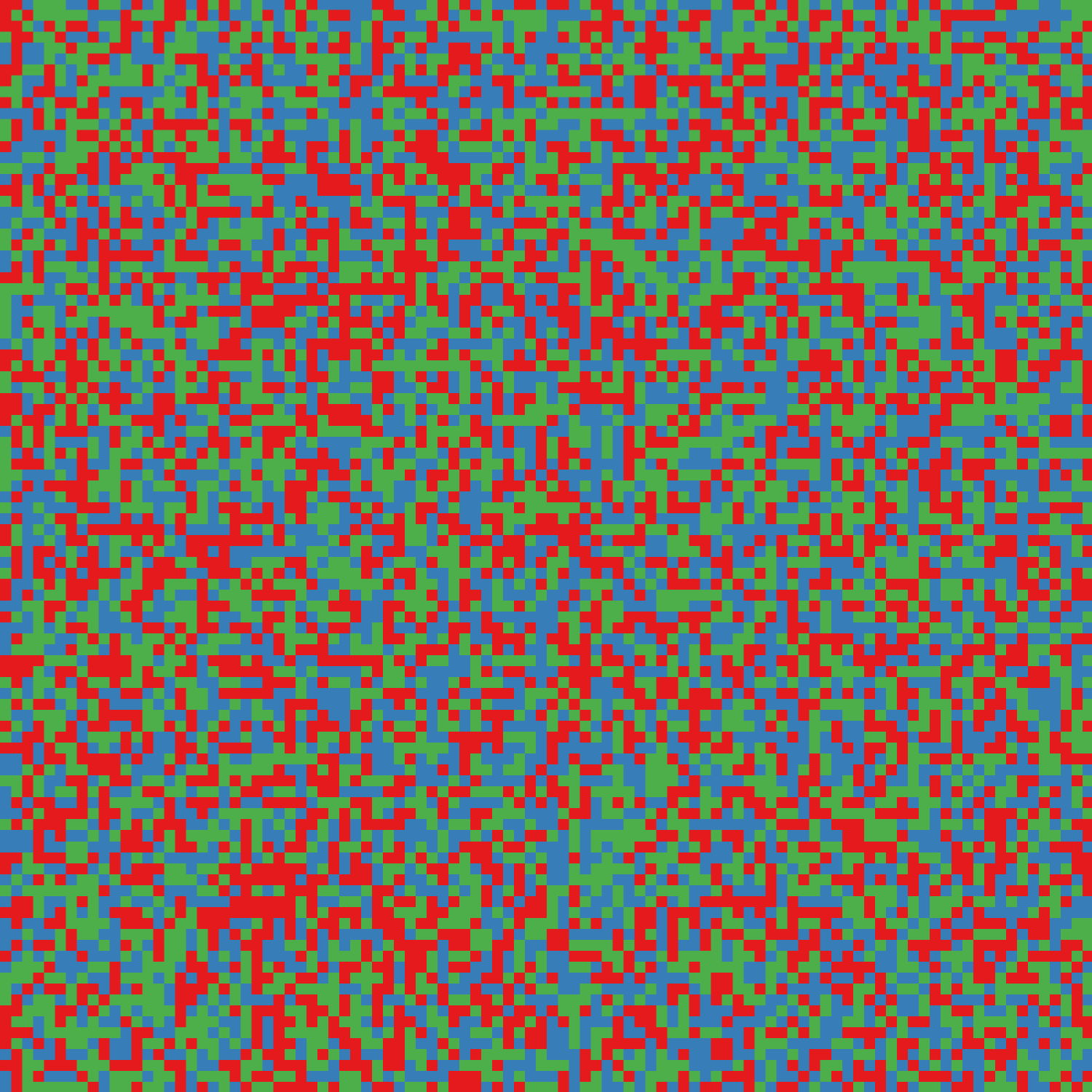, width=0.245\linewidth}}
    {\epsfig{file=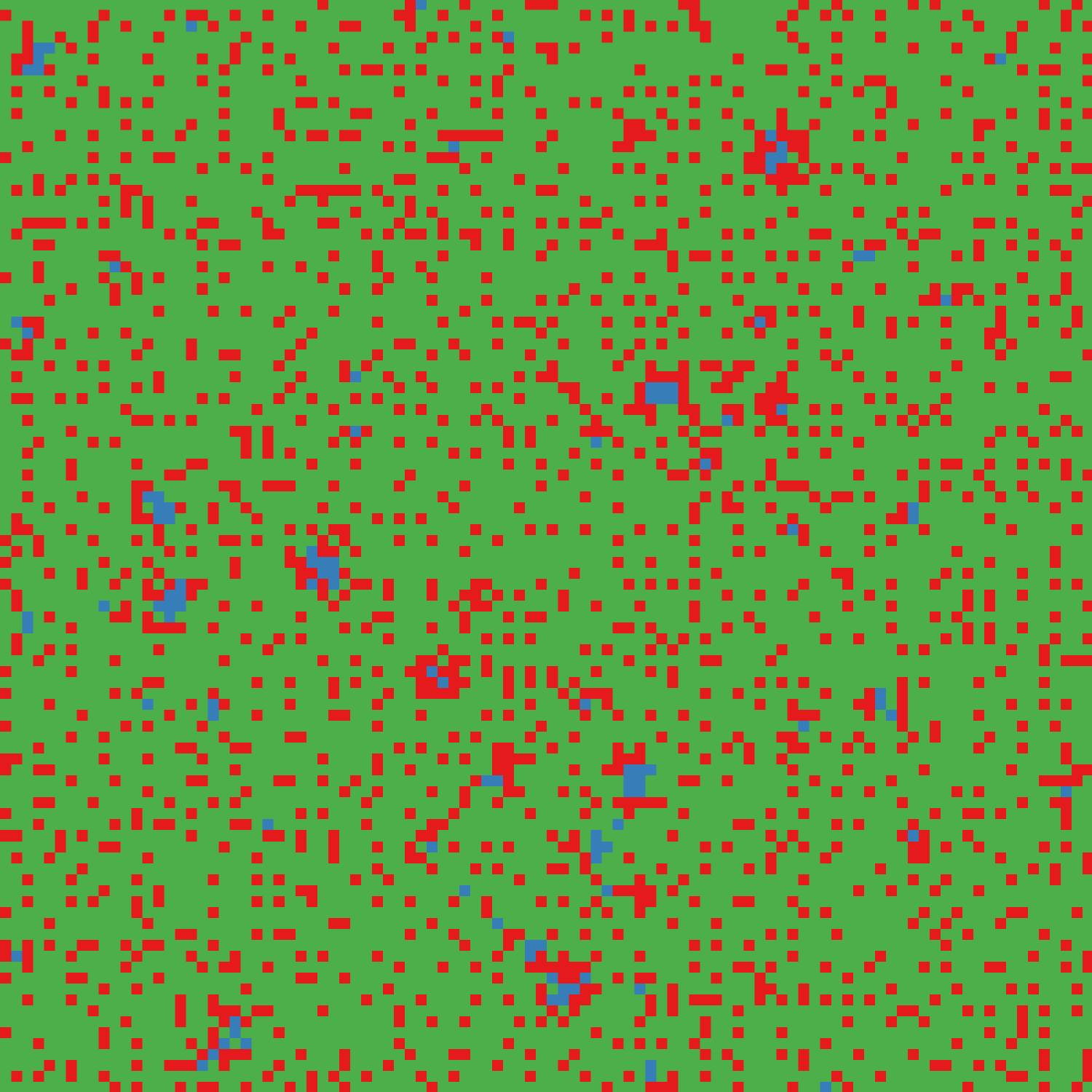, width=0.245\linewidth}}
    {\epsfig{file=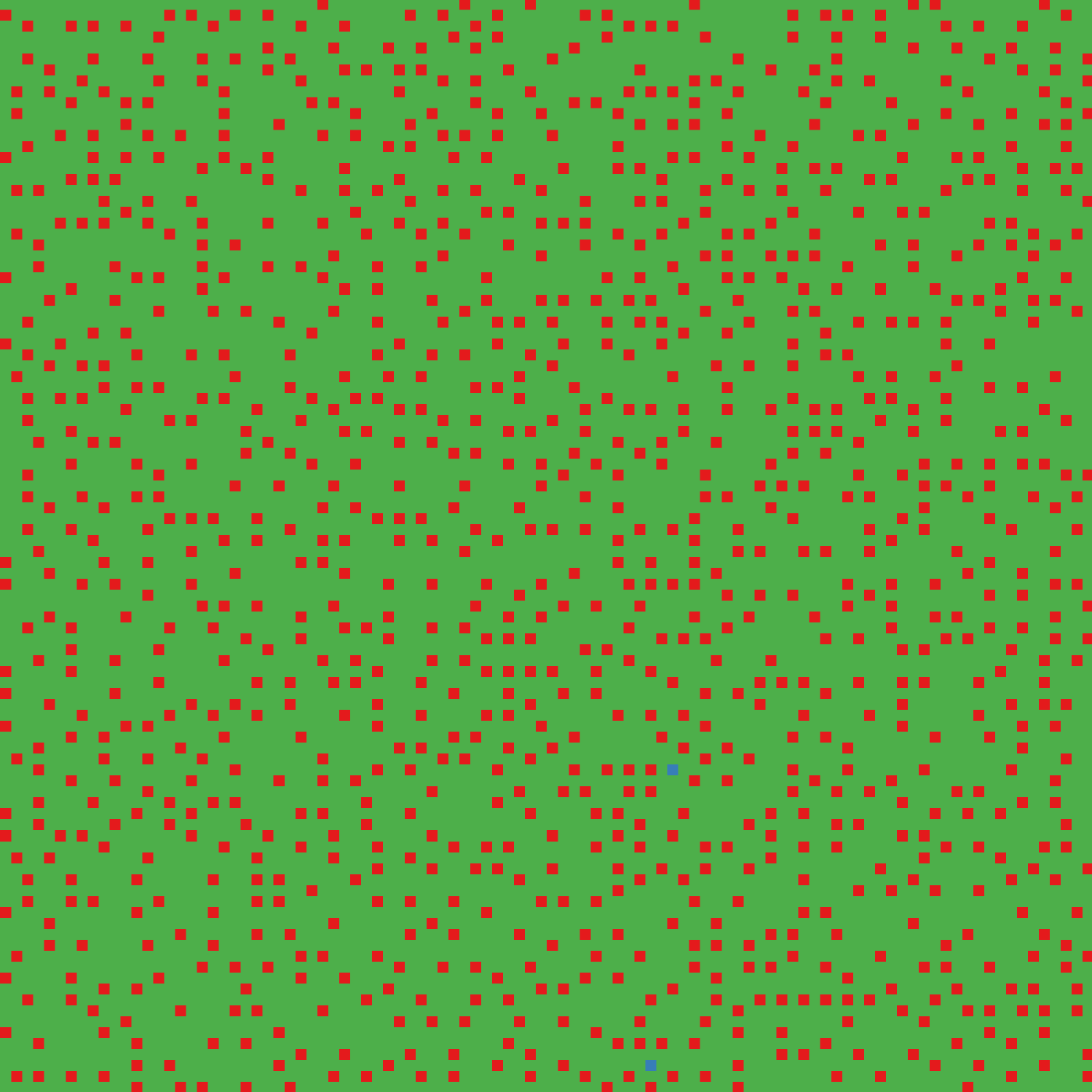, width=0.245\linewidth}}
    {\epsfig{file=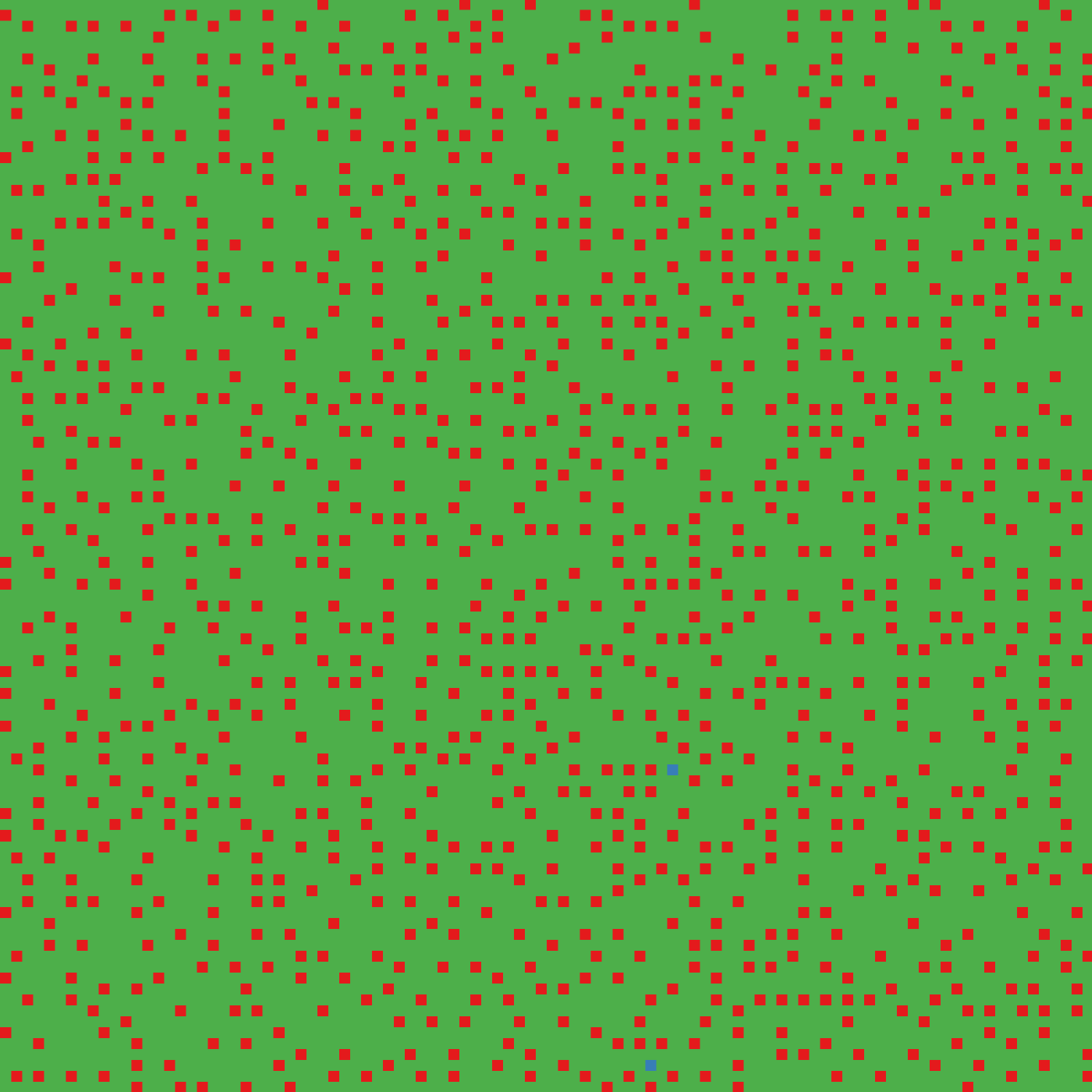, width=0.245\linewidth}}

    \vspace{0.07cm}

    {\epsfig{file=step0, width=0.245\linewidth}}
    {\epsfig{file=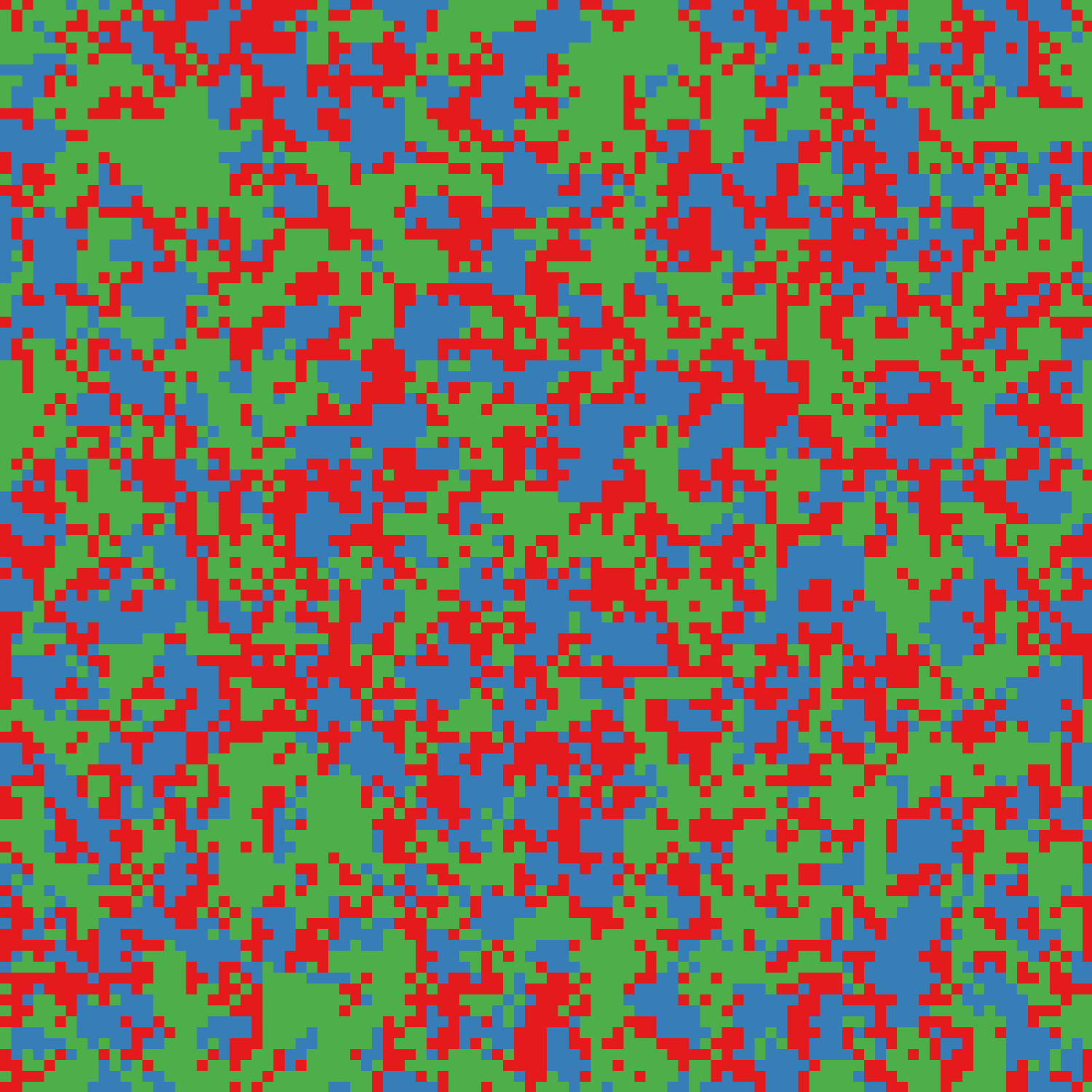, width=0.245\linewidth}}
    {\epsfig{file=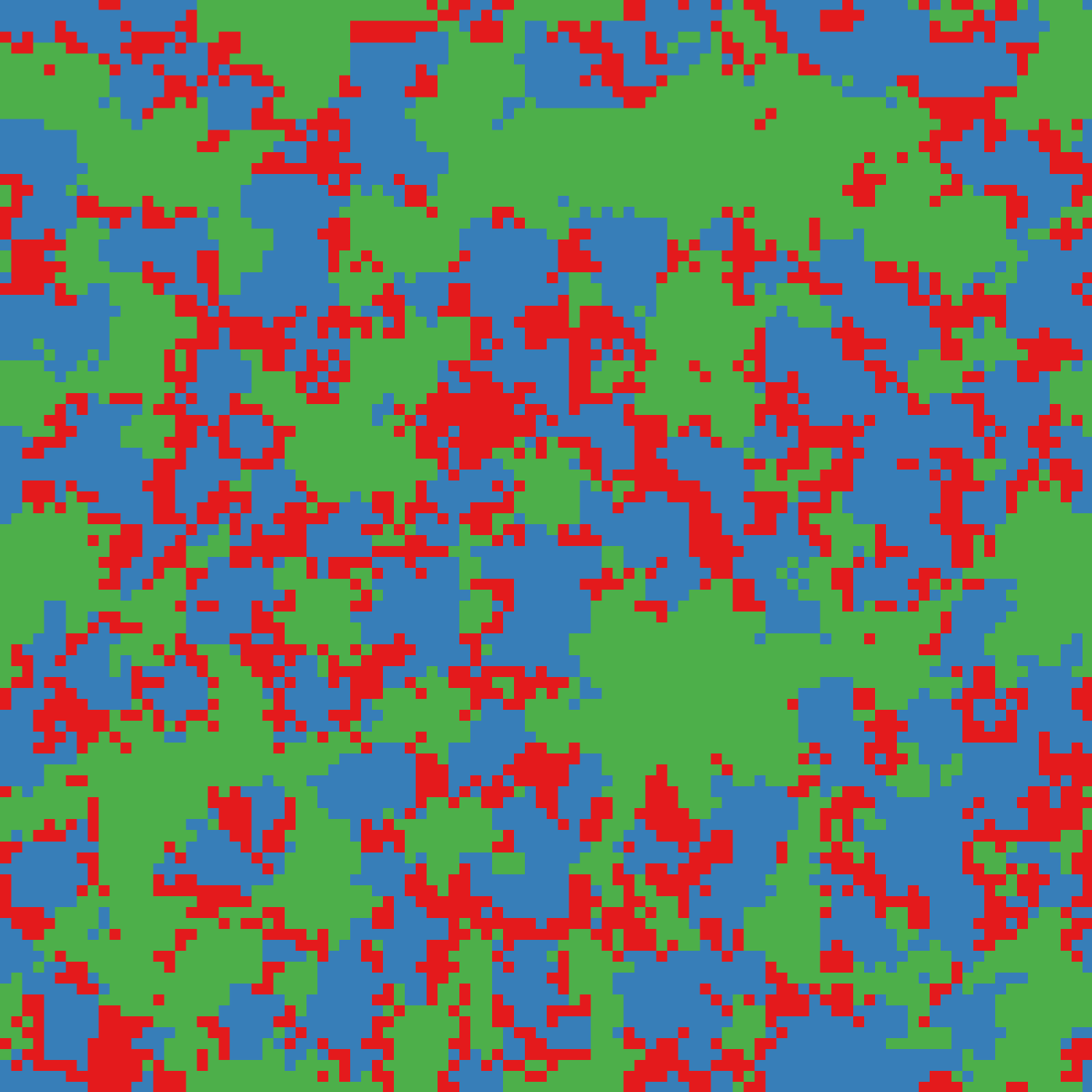, width=0.245\linewidth}}
    {\epsfig{file=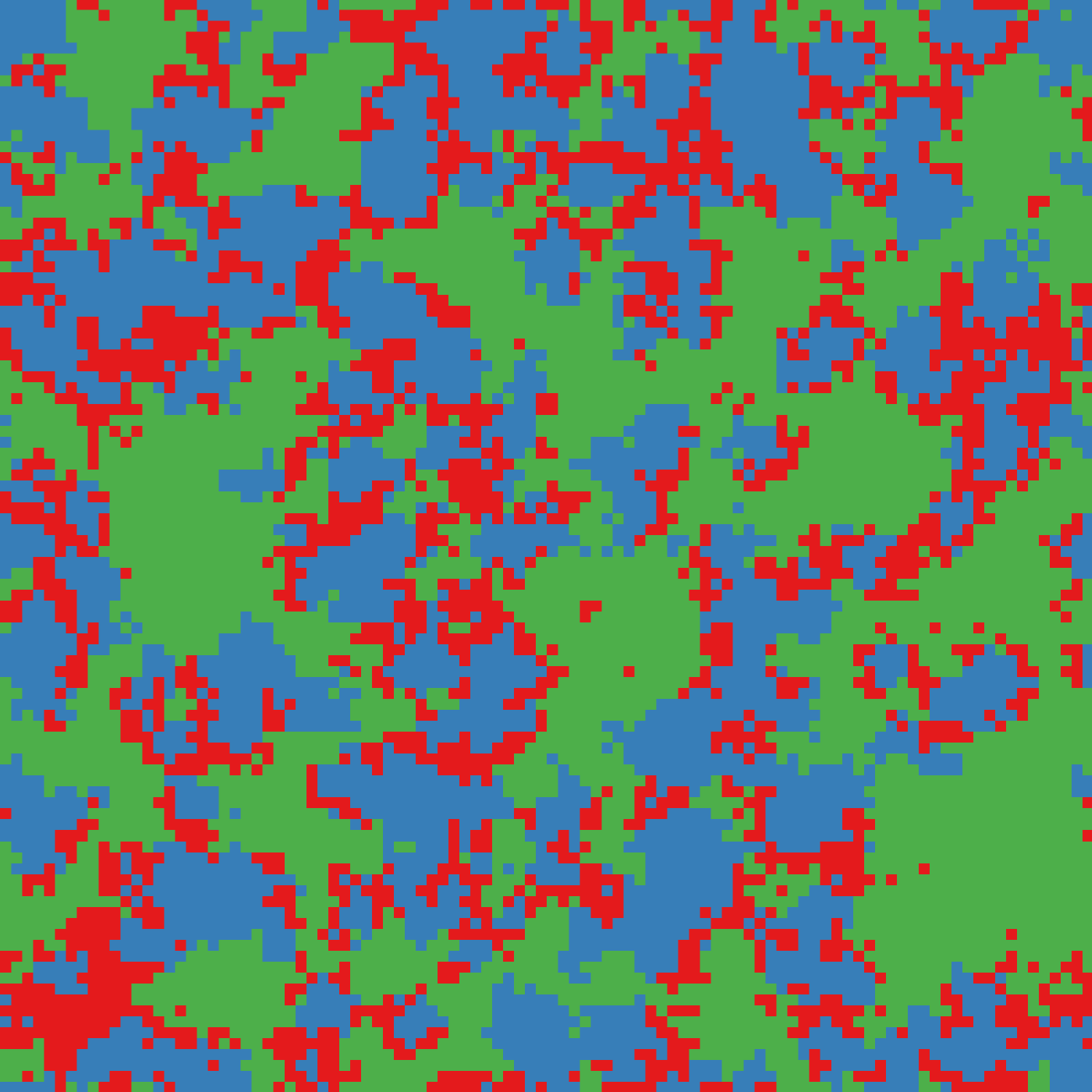, width=0.245\linewidth}}

    \vspace{0.07cm}

    {\epsfig{file=step0, width=0.245\linewidth}}
    {\epsfig{file=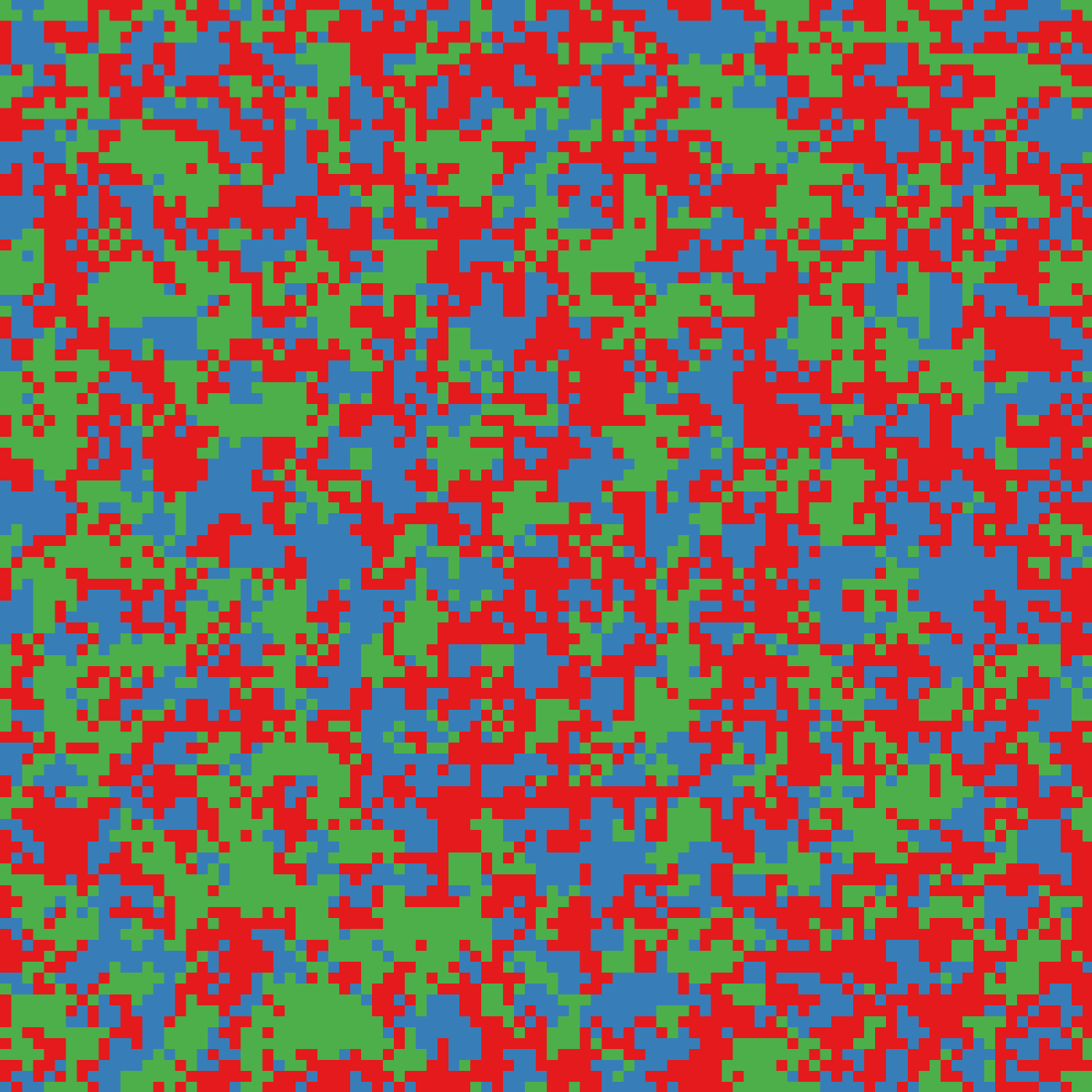, width=0.245\linewidth}}
    {\epsfig{file=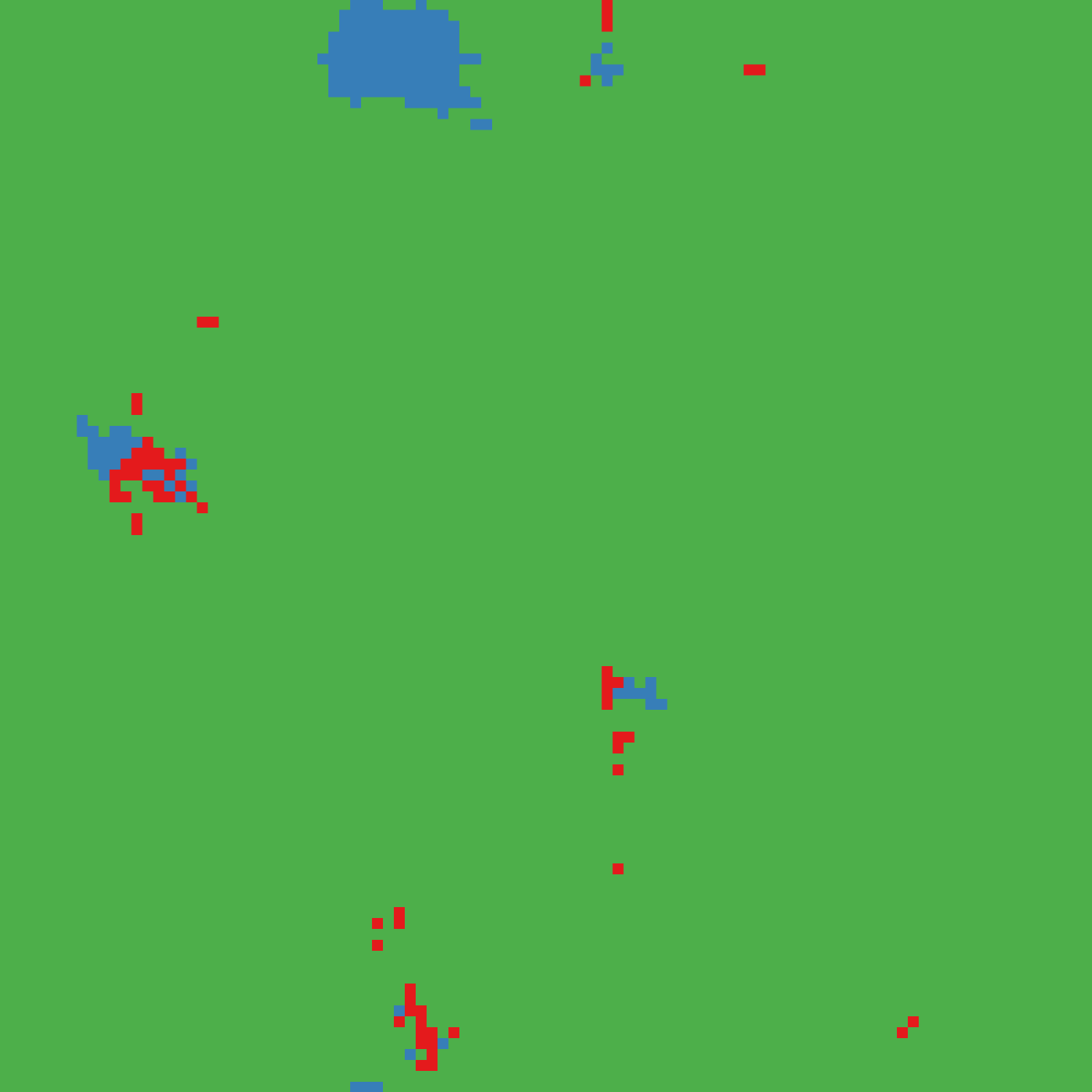, width=0.245\linewidth}}
    {\epsfig{file=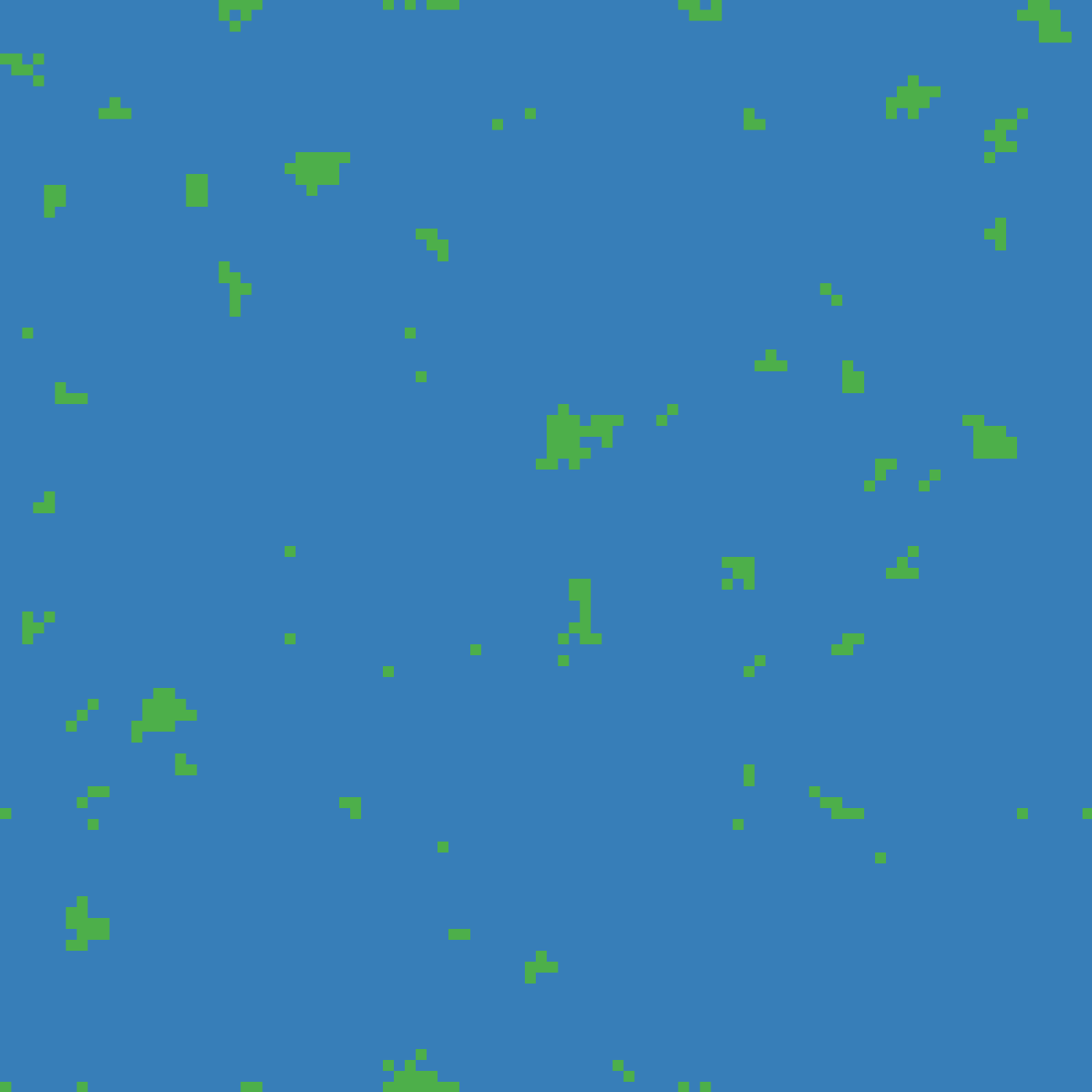, width=0.245\linewidth}}
    \caption{
        Snapshots of the distribution of the strategy in the Monte Carlo steps
        $0$, $45$, $1113$ and $10^5$ (from left to right) for $\Delta/\delta$
        equal to $0.0$, $0.2$ and $1.0$ (from top to bottom). In this Figure,
        cooperators, defectors and abstainers are represented by the colours
        blue, red and green respectively. All results are obtained for $b=1.9$,
        $l=0.6$ and $\delta=0.8$.
    }
    \label{fig:snapshots}
\end{figure*}

In order to further explain the results witnessed in the previous experiments,
we investigate how the population evolves over time.  Figure~\ref{fig:c_mcs}
features the time course of cooperation for three different values of
$\Delta/\delta=\{0.0,\ 0.2,\ 1.0\}$, which are some of the critical points when
$b=1.9$, $l=0.6$ and $\delta=0.8$. Based on these results, in
Figure~\ref{fig:snapshots} we show snapshots for the Monte Carlo steps $0$,
$45$, $1113$ and $10^5$ for the three scenarios shown in
Figure~\ref{fig:c_mcs}.

\begin{figure*}[htb]
    \centering
    {\epsfig{file=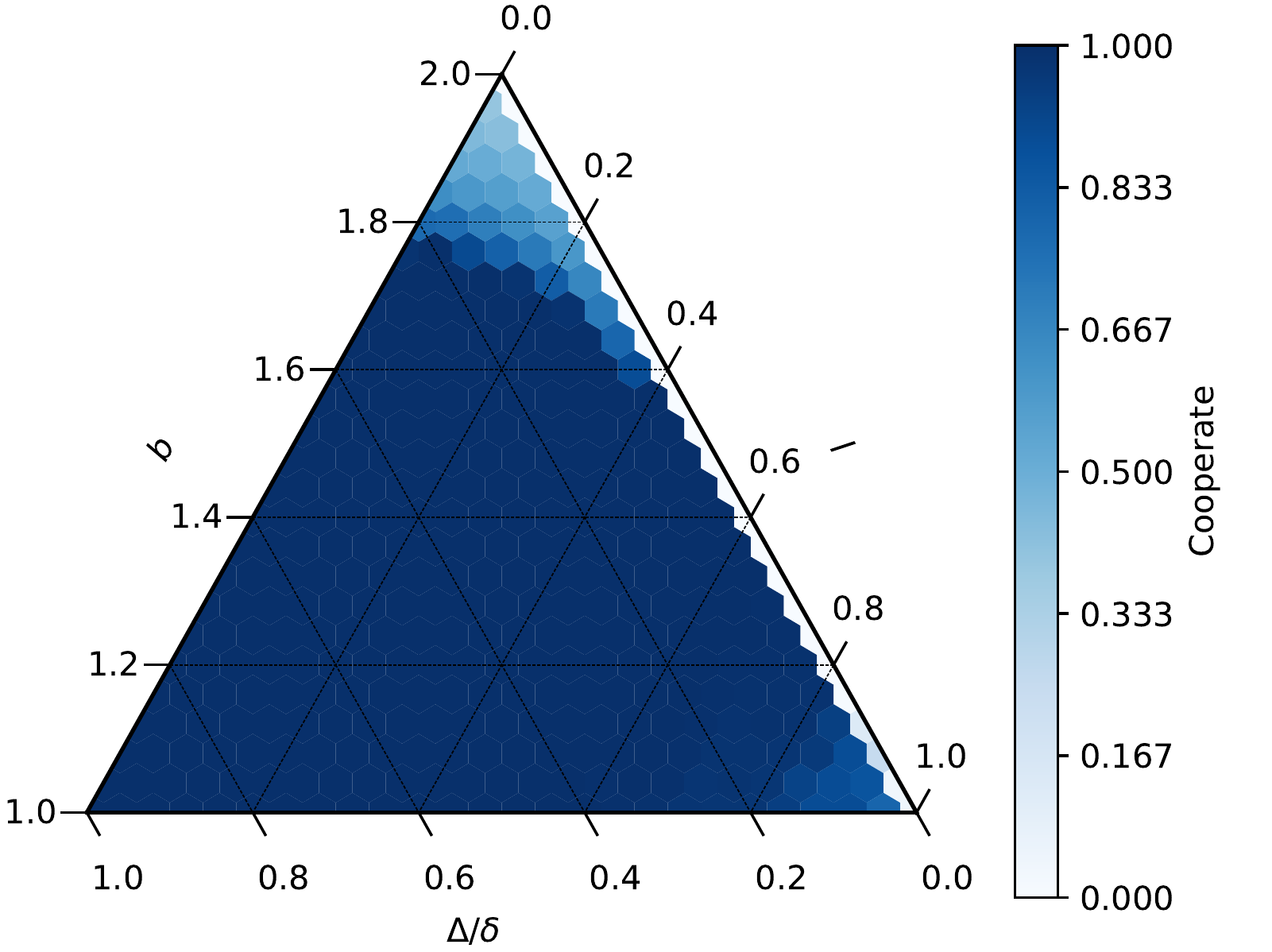, width=0.32\linewidth}}
    {\epsfig{file=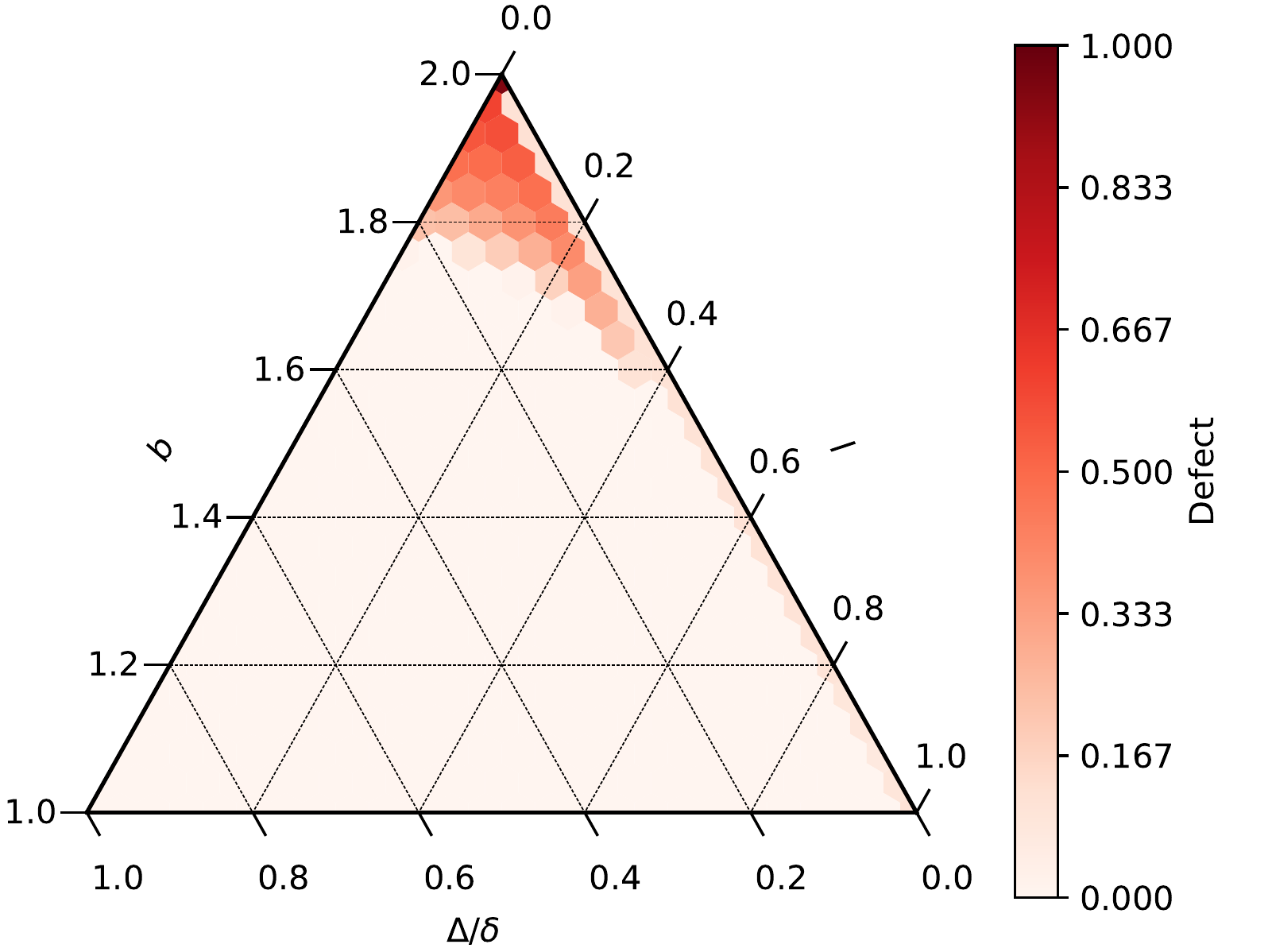, width=0.32\linewidth}}
    {\epsfig{file=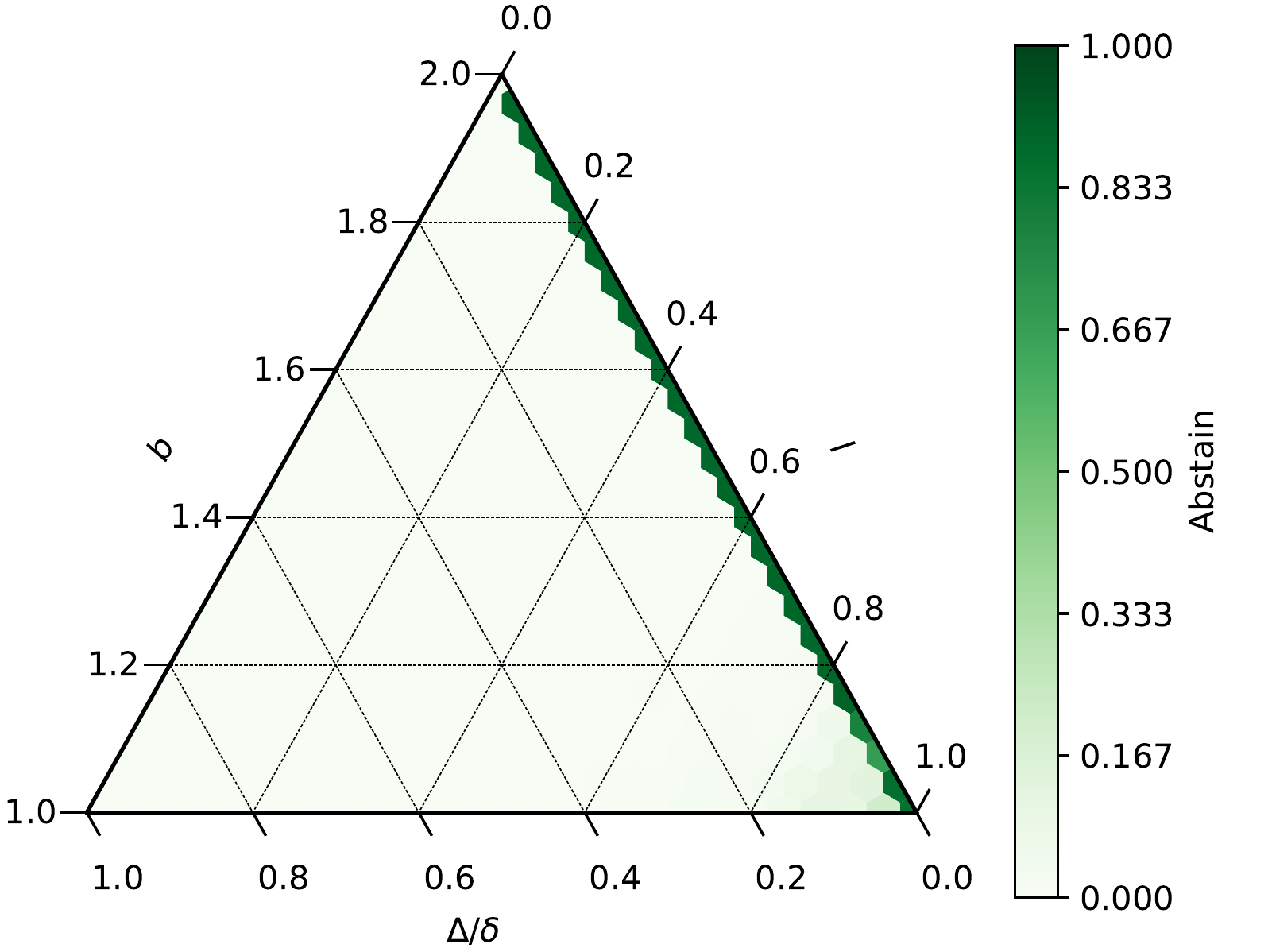, width=0.32\linewidth}}
    \caption{
        Ternary diagrams of different values of $b$, $l$ and $\Delta/\delta$
        for $\delta=0.8$.
    }
    \label{fig:ternary}
\end{figure*}

We see from Figure~\ref{fig:c_mcs} that for the traditional case (i.e.,
$\Delta/\delta=0.0$), abstainers spread quickly and reach a stable state in
which single defectors are completely isolated by abstainers. In this way, as
the payoffs obtained by a defector and an abstainer are the same, neither will
ever change their strategy. In fact, even if a single cooperator survives up to
this stage, for the same aforementioned reason, its strategy will not change
either.

When $\Delta/\delta=0.2$, it is possible to observe some sort of equilibrium
between the three strategies. They reach a state of cyclic competition in which
abstainers invade defectors, defectors invade cooperators and cooperators
invade abstainers.

This behaviour, of balancing the three possible outcomes, is very common in
nature where species with different reproductive strategies remain in
equilibrium in the environment. For instance, the same scenario was observed as
being responsible for preserving biodiversity in the neighbourhoods of the
\textit{Escherichia coli}, which is a bacteria commonly found in the lower
intestine of warm-blooded organisms. According to \citet{Fisher2008}, studies
were performed with three natural populations mixed together, in which one
population produces a natural antibiotic but is immune to its effects; a second
population is sensitive to the antibiotic but can grow faster than the third
population; and the third population is resistant to the antibiotic.

Because of this balance, they observed that each population ends up
establishing its own territory in the environment, as the first population
could kill off any other bacteria sensitive to the antibiotic, the second
population could use their faster growth rate to displace the bacteria which
are resistant to the antibiotic, and the third population could use their
immunity to displace the first population.

Another interesting behaviour is noticed for $\Delta/\delta=1.0$. In this
scenario, defectors are dominated by abstainers, allowing a few clusters of
cooperators to survive. As a result of the absence of defectors, cooperators
invade abstainers and dominate the environment.

\subsection{Investigating the Relationship between $\Delta/\delta$, $b$ and $l$}

To investigate the outcomes in other scenarios, we explore a wider range of
settings by varying the values of the temptation to defect ($b$), the loner's
payoff ($l$) and the link weight amplitude ($\Delta/\delta$) for a fixed
value of weight heterogeneity ($\delta = 0.8$).

As shown in Figure~\ref{fig:ternary}, cooperation is the dominant strategy in
the majority of cases. Note that in the traditional case, with an unweighted
and static network, i.e., $\Delta/\delta=0.0$, abstainers dominate in all
scenarios illustrated in this ternary diagram. In addition, it is also possible
to observe that certain combinations of $l$, $b$ and $\Delta/\delta$ guarantee
higher levels of cooperation.  In these scenarios, cooperators are protected by
abstainers against exploitation from defectors. In most cases, for populations
with the loner's payoff, $l=[0.4, 0.8]$, cooperation is promoted the most.

Although the combinations shown in Figure~\ref{fig:ternary} for higher values
of b ($b>1.8$) are just a small subset of an infinite number of possible
values, it is clearly shown that a reasonable fraction of cooperators can
survive even in an extremely adverse situation where the advantage of defecting
is very high.  Indeed, our results show that some combinations of high values
of $l$ and $\Delta/\delta$ such as for $\Delta/\delta=1.0$ and $l=0.6$, can
further improve the levels of cooperation, allowing for the full dominance of
cooperation.

In summary, we see that the use of a coevolutionary model in the Optional
Prisoner's Dilemma game allows for the emergence of cooperation.

\section{\uppercase{Conclusions and Future Work}}
\label{sec:conclusion}
\noindent
In this paper, we studied the impact of abstinence in the Prisoner's Dilemma
game using a coevolutionary spatial model in which both game strategies and
link weights between agents evolve over time. We considered a population of
agents who were initially organised on a lattice grid where agents can only
play with their eight immediate neighbours. Using a Monte Carlo simulation
approach, a number of experiments were performed to observe the emergence of
cooperation, defection and abstinence in this environment. At each Monte Carlo
time step, an agent's strategy (cooperate, defect, abstain) and the link weight
between agents can be updated.

The payoff received by an agent after playing with another agent is a product
of the strategy played and the weight of the link between agents. We explored
the effect of the link update rules by varying the values of the link weight
amplitude $\Delta/\delta$, the loner's payoff $l$, and the temptation to defect
$b$. The aims were to understand the relationship between these parameters
and also investigate the evolution of cooperation when abstainers are present
in the population.

Results showed that, in adverse scenarios where $b$ is very high
(i.e.~$b\ge1.9$), some combinations of high values of $l$ and $\Delta/\delta$,
such as for $\Delta/\delta=1.0$ and $l=0.6$, can further improve the levels of
cooperation, even resulting in the full dominance of cooperation.

When $\Delta/\delta=0.2$, $b=1.9$ and $\delta=0.8$, it was possible to observe
a balance among the three strategies, indicating that, for some parameter
settings, the Optional Prisoner's Dilemma game is intransitive. In other words,
such scenarios produce a loop of dominance in which abstainer agents beat
defector agents, defector agents beat cooperator agents and cooperator agents
beat abstainer agents.

In summary, the difference between the outcomes of $\Delta/\delta=0.0$ (i.e., a
static environment with unweighted links) and $\Delta/\delta=(0.0,\ 1.0]$
clearly showed that the coevolutionary model is very advantageous to the
promotion of cooperative behaviour. In most cases, with populations with the
loner's payoff, ${l=[0.4,\ 0.8]}$, cooperation is promoted the most. Moreover,
results also showed that cooperators are protected by abstainers against
exploitation from defectors.

Although recent research has considered coevolving game strategy and link
weights (Section~\ref{sec:related}), to our knowledge the investigation of such
a coevolutionary model with optional games has not been studied to date. We
conclude that the combination of both of these trends in evolutionary game
theory may shed additional light on gaining an in-depth understanding of the
emergence of cooperative behaviour in real-world scenarios.

Future work will consider the exploration of different topologies and the
influence of a  wider range of scenarios, where, for example, agents could
rewire their links, which, in turn, adds another level of complexity to the
model. Future work will also involve applying our studies and results to model
realistic scenarios, such as social networks and real biological networks.

\section*{\uppercase{Acknowledgements}}
\noindent This work was supported by the National Council for Scientific and Technological Development (CNPq-Brazil).

\bibliographystyle{apalike}
{\small \bibliography{ecta16}}

\begin{thebibliography}{}

\bibitem[Batali and Kitcher, 1995]{Batali1995}
Batali, J. and Kitcher, P. (1995).
\newblock Evolution of altruism in optional and compulsory games.
\newblock {\em Journal of Theoretical Biology}, 175(2):161--171.

\bibitem[Cao et~al., 2011]{Cao2011}
Cao, L., Ohtsuki, H., Wang, B., and Aihara, K. (2011).
\newblock Evolution of cooperation on adaptively weighted networks.
\newblock {\em Journal of Theoretical Biology}, 272(1):8 -- 15.

\bibitem[Cardinot et~al., 2016]{Cardinot2016}
Cardinot, M., Gibbons, M., O'Riordan, C., and Griffith, J. (2016).
\newblock Simulation of an optional strategy in the prisoner's dilemma in
  spatial and non-spatial environments.
\newblock In {\em From Animals to Animats 14 (SAB 2016)}, pages 145--156, Cham.
  Springer International Publishing.

\bibitem[Chen and Wang, 2008]{Chen2008}
Chen, X. and Wang, L. (2008).
\newblock Promotion of cooperation induced by appropriate payoff aspirations in
  a small-world networked game.
\newblock {\em Physical Review E}, 77:017103.

\bibitem[Fisher, 2008]{Fisher2008}
Fisher, L. (2008).
\newblock {\em Rock, Paper, Scissors: Game Theory in Everyday Life}.
\newblock Basic Books.

\bibitem[Fu et~al., 2007]{Fu2007}
Fu, F., Liu, L.-H., and Wang, L. (2007).
\newblock Evolutionary prisoner's dilemma on heterogeneous newman-watts
  small-world network.
\newblock {\em The European Physical Journal B}, 56(4):367--372.

\bibitem[Ghang and Nowak, 2015]{Ghang2015}
Ghang, W. and Nowak, M.~A. (2015).
\newblock {Indirect reciprocity with optional interactions}.
\newblock {\em Journal of Theoretical Biology}, 365:1--11.

\bibitem[Gómez-Gardeñes et~al., 2011]{Gomez2011}
Gómez-Gardeñes, J., Romance, M., Criado, R., Vilone, D., and Sánchez, A.
  (2011).
\newblock Evolutionary games defined at the network mesoscale: The public goods
  game.
\newblock {\em Chaos}, 21(1).

\bibitem[Hauert et~al., 2008]{Hauert2008}
Hauert, C., Traulsen, A., Brandt, H., and Nowak, M.~A. (2008).
\newblock {Public goods with punishment and abstaining in finite and infinite
  populations}.
\newblock {\em Biological Theory}, 3(2):114--122.

\bibitem[Huang et~al., 2015]{Huang2015}
Huang, K., Zheng, X., Li, Z., and Yang, Y. (2015).
\newblock Understanding cooperative behavior based on the coevolution of game
  strategy and link weight.
\newblock {\em Scientific Reports}, 5:14783.

\bibitem[Jeong et~al., 2014]{Jeong2014}
Jeong, H.-C., Oh, S.-Y., Allen, B., and Nowak, M.~A. (2014).
\newblock Optional games on cycles and complete graphs.
\newblock {\em Journal of Theoretical Biology}, 356:98--112.

\bibitem[Nowak and May, 1992]{Nowak1992}
Nowak, M.~A. and May, R.~M. (1992).
\newblock Evolutionary games and spatial chaos.
\newblock {\em Nature}, 359(6398):826--829.

\bibitem[Olejarz et~al., 2015]{Olejarz2015}
Olejarz, J., Ghang, W., and Nowak, M.~A. (2015).
\newblock {Indirect Reciprocity with Optional Interactions and Private
  Information}.
\newblock {\em Games}, 6(4):438--457.

\bibitem[Perc and Szolnoki, 2010]{Perc2010}
Perc, M. and Szolnoki, A. (2010).
\newblock {Coevolutionary games -- A mini review}.
\newblock {\em Biosystems}, 99(2):109--125.

\bibitem[Szab\'o and Hauert, 2002]{Hauert2002}
Szab\'o, G. and Hauert, C. (2002).
\newblock Evolutionary prisoner's dilemma games with voluntary participation.
\newblock {\em Physical Review E}, 66:062903.

\bibitem[Szolnoki and Perc, 2009]{Szolnoki2009}
Szolnoki, A. and Perc, M. (2009).
\newblock Promoting cooperation in social dilemmas via simple coevolutionary
  rules.
\newblock {\em The European Physical Journal B}, 67(3):337--344.

\bibitem[Szolnoki and Perc, 2016]{Szolnoki2016}
Szolnoki, A. and Perc, M. (2016).
\newblock Leaders should not be conformists in evolutionary social dilemmas.
\newblock {\em Scientific Reports}, 6:23633.

\bibitem[Wang et~al., 2014]{Wang2014}
Wang, Z., Szolnoki, A., and Perc, M. (2014).
\newblock Self-organization towards optimally interdependent networks by means
  of coevolution.
\newblock {\em New Journal of Physics}, 16(3):033041.

\bibitem[Xia et~al., 2015]{Xia2015}
Xia, C.-Y., Meloni, S., Perc, M., and Moreno, Y. (2015).
\newblock Dynamic instability of cooperation due to diverse activity patterns
  in evolutionary social dilemmas.
\newblock {\em EPL}, 109(5):58002.

\bibitem[Zimmermann et~al., 2001]{Zimmermann2001}
Zimmermann, M.~G., Egu{\'i}luz, V.~M., and Miguel, M.~S. (2001).
\newblock {\em {Cooperation, Adaptation and the Emergence of Leadership}},
  pages 73--86.
\newblock Springer, Berlin, Heidelberg.

\bibitem[Zimmermann et~al., 2004]{Zimmermann2004}
Zimmermann, M.~G., Egu\'{\i}luz, V.~M., and San~Miguel, M. (2004).
\newblock Coevolution of dynamical states and interactions in dynamic networks.
\newblock {\em Physical Review E}, 69:065102.

\end{thebibliography}

\end{document}